\documentclass[10pt,twocolumn,letterpaper]{article}
\usepackage[pagenumbers]{arxivstype}

\usepackage{graphicx}
\usepackage{afterpage}
\usepackage{placeins}
\usepackage{multirow}
\usepackage{colortbl,xcolor}
\usepackage{float}
\usepackage{amsmath}
\usepackage{amssymb}
\usepackage{wrapfig}
\usepackage{tabularx}
\usepackage{cuted} %

\definecolor{tablebg}{rgb}{0.95,0.95,0.98}
\definecolor{sota}{rgb}{1,0.5,0.5}
\definecolor{secondbest}{rgb}{1,0.75,0.5}

\newcommand{\ourmodel}{\textnormal{\textsc{ScaleWorld}}}

\definecolor{linkblue}{rgb}{0.21,0.49,0.74}
\usepackage[pagebackref,breaklinks,colorlinks,allcolors=linkblue]{hyperref}

\usepackage{url}

\title{ScaleWorld: High-Resolution World Generation through Shared Prediction and Refinement}

\author{
Yansong Qu$^{1,*}$ \quad Wesker Zheng$^{2,*}$ \quad Quanjian Song$^{1,*}$ \quad Zhiyuan Zhang$^{3}$\\
Xiao Cao$^{4}$ \quad Shengchuan Zhang$^{1}$ \quad Liujuan Cao$^{1}$\\[3pt]
{\normalsize $^{1}$Xiamen University \quad $^{2}$Southeast University}\\
{\normalsize $^{3}$City University of Hong Kong \quad $^{4}$National University of Singapore}\\[3pt]
{\small\url{https://quyans.github.io/ScaleWorld/}}
}

\begin{document}
\maketitle
\begingroup
\renewcommand{\thefootnote}{\fnsymbol{footnote}}
\footnotetext[1]{Equal contribution.}
\endgroup
\begin{strip}
    \vspace{-36pt}
    \centering
    \captionsetup{hypcap=false}
    \includegraphics[width=\linewidth]{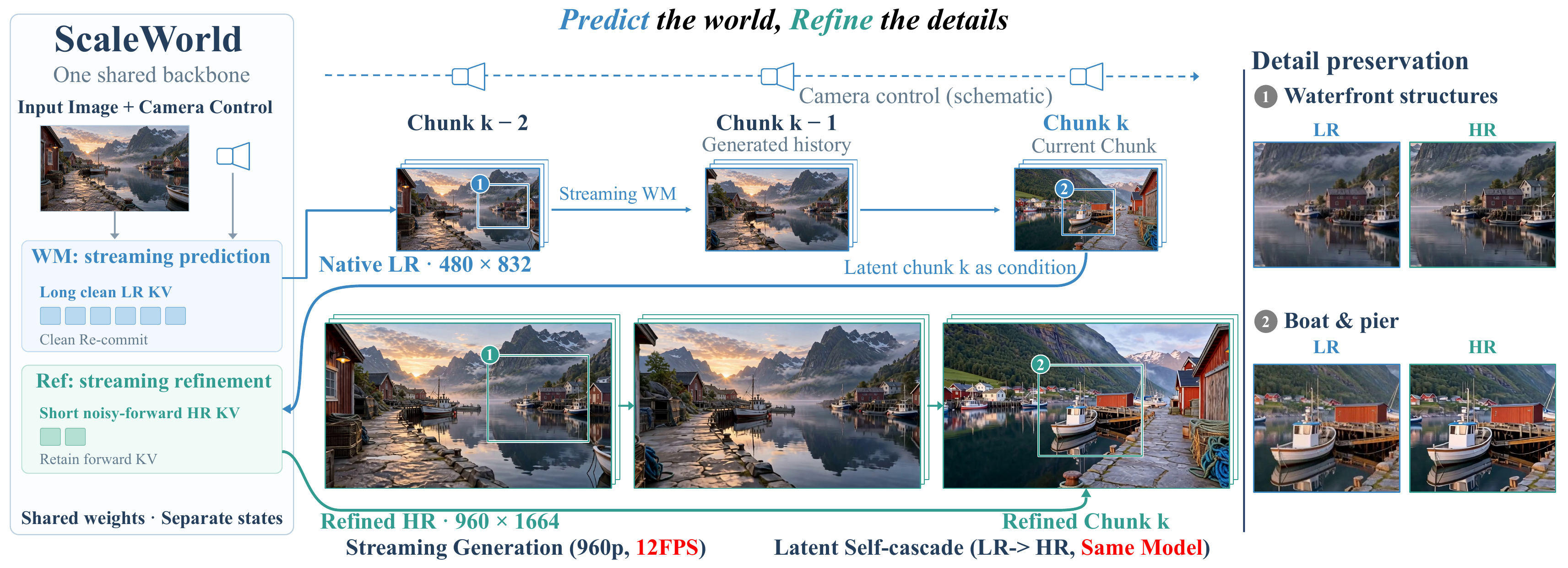}
    \captionof{figure}{\textbf{\ourmodel{} uses one shared backbone for streaming
    world prediction and high-resolution refinement.}
    Each completed low-resolution chunk conditions its high-resolution
    counterpart, while separate causal states retain a long clean
    prediction history and a short noisy-forward refinement history.
    Matched crops highlight the recovered structures and textures.}
    \label{fig:teaser}
    \captionsetup{hypcap=true}
\end{strip}

\begin{abstract}

High-resolution generation enables video world models to render fine-grained scene structures and textures, but directly generating high-resolution video incurs substantial computational and memory costs.
We present \ourmodel{}, a causal video model that unifies low-resolution world prediction and high-resolution refinement within a single shared backbone.
By refining its own predictions, \ourmodel{} produces high-resolution world videos without requiring a separate refinement model.
For each video chunk, the shared backbone first completes a low-resolution prediction and then performs one-step high-resolution refinement conditioned on that prediction.
To support efficient streaming generation, we maintain separate causal states with a long low-resolution history for prediction and a short high-resolution history for refinement.
To address task interference during distillation, we separate specialist optimization from capability consolidation and introduce task-asymmetric on-policy distillation (TA-OPD).
This procedure transfers prediction and refinement behaviors from specialized teachers into a shared student through local velocity matching under their respective sampling and history policies.
{Extensive experiments demonstrate that \ourmodel{} achieves high-quality world generation with improved computational and memory efficiency.}

\end{abstract}

\section{Introduction}
\label{sec:intro}

Video world models generate evolving visual environments conditioned on
observations and control inputs
\citep{sun2025worldplay,gao2026infiniteworlds,qian2026matrixgame35}.
High-resolution generation enables these models to depict fine-grained
scene structures and textures as the environment evolves.
Causal video diffusion supports incremental generation
\citep{yin2025causvid,huang2025selfforcing}, but directly generating
high-resolution video incurs substantial computational and memory costs.
Increasing spatial resolution raises both denoising computation and
temporal-state storage, making high-resolution output costly to sustain
in streaming world generation.

Coarse-to-fine pipelines address this cost by predicting at low resolution
and refining the result with a dedicated model
\citep{luxury2026ultraflash,zhu2026sanawm}.
This reduces the spatial cost of world prediction, but deploying separate
predictors and refiners still requires storing two sets of model weights.
Prediction and refinement share visual priors for structure, appearance,
and motion, motivating a shared backbone.
Cross-resolution reuse has been explored by Self-Cascade, Pyramid Flow,
and HiStream \citep{guo2024selfcascade,jin2024pyramidflow,qiu2025histream}.
Our focus is the coexistence of independently usable world prediction
and conditional streaming refinement within one causal model.
{We therefore ask: \emph{can a world model learn high-resolution
refinement while retaining native prediction?}}

We present \ourmodel{}, a shared causal model for streaming world
prediction and streaming high-resolution refinement.
Both roles follow a chunk-wise autoregressive paradigm: the model first
completes a low-resolution (LR) prediction, then uses the same backbone
to perform one-step high-resolution (HR) refinement conditioned on that
prediction and its own refinement history.
No future LR chunks are required, so each completed prediction can be
refined without waiting for a full video.
The native LR output remains independently usable, and HR outputs do not
feed back into prediction history.
{This design must accommodate the roles' distinct temporal requirements
while consolidating both capabilities in shared parameters.}

To support efficient inference, we share model weights while maintaining
separate causal states.
World prediction retains a longer clean history at low resolution to
infer scene evolution.
Refinement already receives the current LR structure, content, and motion,
and uses a shorter HR history to support the continuity of synthesized detail.
It further retains key--value (KV) states from the denoising forward pass
as \emph{noisy-forward history}, avoiding an additional HR model evaluation
for history construction.
These asymmetric histories limit the cost of refinement while preserving
longer temporal context for world prediction.

Sharing visual representations does not ensure that prediction and
refinement can be effectively distilled together.
In our joint distillation experiments, we observe conflicting gradients
between the two tasks, and {better reconstruction during refinement} can accompany
reduced temporal stability in world prediction
(Appendix~\ref{sec:supp_gradient_conflict}).
We therefore separate specialist optimization from capability consolidation.
Starting from a jointly trained and causally adapted model, we first
optimize prediction and refinement specialists under their respective
sampling and history policies.
Building on on-policy distillation \citep{li2026diffusionopd,zhou2026danceopd},
we introduce \emph{task-asymmetric on-policy distillation} (TA-OPD) to
consolidate the frozen specialists into a shared student.
TA-OPD matches expert velocities at student-visited denoising queries for
prediction, and at matched conditional noise queries with network-specific
recurrent histories for refinement, respecting the roles' different execution policies.

We evaluate \ourmodel{} on camera-controlled world generation, examining
native prediction retention, refinement of fixed generated inputs, and
complete HR generation.
The shared model retains competitive native prediction performance and
achieves HR generation quality comparable to a separate specialist cascade.
Matched deployments of the same unified checkpoint show that sharing
weights reduces peak GPU memory while preserving throughput.
{Noisy-forward history accelerates refinement over clean recommit
by avoiding an extra HR forward pass.}

Our contributions are threefold:
\begin{itemize}
    \item {A shared causal world model for streaming prediction and streaming
    HR refinement, refining each completed LR prediction with the same
    backbone while retaining an independently usable native prediction path.}
    \item {A training and inference framework combining TA-OPD and asymmetric
    causal histories to meet the distinct optimization and temporal
    requirements of both roles.}
    \item Empirical evidence of competitive prediction and HR generation
    quality, with controlled deployment measurements of memory savings
    from weight sharing and timing comparisons of refinement-history construction.
\end{itemize}

\section{Related Work}
\label{sec:relatedwork}
\label{sec:related_world}

\textbf{Controllable world generation.}
Video diffusion models learn rich appearance and motion priors
\citep{ho2022videodiffusion,wan2025}.
Genie~\citep{bruce2024genie} learns action-controllable environments
from unlabeled videos, while GameNGen~\citep{valevski2025gamengen}
simulates interactive gameplay through action-conditioned frame
prediction.
Beyond games, GAIA-1~\citep{hu2023gaia1} generates driving scenarios
conditioned on video, text, and actions.
Cosmos~\citep{nvidia2025cosmos} provides pretrained diffusion and
autoregressive video world foundation models for physical AI.

For efficient streaming generation, CausVid~\citep{yin2025causvid}
and Self Forcing~\citep{huang2025selfforcing} combine few-step
generation with autoregressive execution.
Causal Forcing~\citep{zhu2026causalforcing} uses autoregressive
teacher trajectories for initialization before distribution matching
distillation.
WorldPlay~\citep{sun2025worldplay},
LingBot-World 2.0~\citep{gao2026infiniteworlds}, and
Matrix-Game 3.5~\citep{qian2026matrixgame35} advance long-horizon
controlled world generation.
SANA-WM~\citep{zhu2026sanawm} combines camera-controlled generation
with a separate long-video refiner.
Our work focuses on high-resolution generation within a shared
world-model backbone, with streaming execution for both prediction
and refinement.

\textbf{Shared coarse-to-fine generation.}
Self-Cascade~\citep{guo2024selfcascade} reuses a pretrained
low-resolution diffusion model for higher-resolution adaptation.
Pyramid Flow~\citep{jin2024pyramidflow} jointly learns resolution
stages within a unified flow, while
PyramidalWan~\citep{korzhenkov2026pyramidalwan} adapts pretrained
video models to pyramidal generation.
HiStream~\citep{qiu2025histream} performs early denoising at low
resolution and later denoising at high resolution with dual-resolution
caches. It also downsamples the final high-resolution output to refresh
the low-resolution history cache.
In \ourmodel{}, each prediction chunk completes its native output
before conditional refinement. The two roles maintain separate causal
states, and refined outputs do not feed back into prediction.

Other forms of parameter reuse include
LongCat-Video~\citep{meituan2025longcat}, which trains a refinement
expert through LoRA adaptation of its base model, and
HiFI~\citep{hur2025hifi}, which shares one model across frame
interpolation and spatial upsampling.
Our study examines parameter sharing between world prediction and
conditional refinement under distinct sampling budgets and
causal-memory policies.

\textbf{Generative video refinement.}
VEnhancer~\citep{he2024venhancer} improves generated videos through
joint spatial and temporal enhancement.
SeedVR~\citep{wang2025seedvr} studies diffusion-based video
restoration, while SeedVR2~\citep{wang2026seedvr2} enables one-step
restoration through adversarial post-training.
FlashVSR~\citep{zhuang2025flashvsr} targets one-step streaming
super-resolution.
Ultra Flash~\citep{luxury2026ultraflash} combines an autoregressive
low-resolution generator with a dedicated causal refiner, using latent
upsampling, single-step distillation, and preference optimization
on generated contexts.
Building on advances in streaming refinement, \ourmodel{} consolidates
refinement and native world prediction within one shared backbone.

\textbf{Distillation and capability composition.}
Distribution matching distillation (DMD)~\citep{yin2024dmd}
transfers a diffusion model's generative behavior to a one-step student.
Self Forcing~\citep{huang2025selfforcing} addresses the exposure bias
of autoregressive video generation through student-rollout training.
DiffusionOPD~\citep{li2026diffusionopd} consolidates specialist
capabilities on student trajectories, while
DanceOPD~\citep{zhou2026danceopd} composes capabilities through
routed velocity-field matching.
For autoregressive video, OPSD-V~\citep{liu2026opsdv} supervises
student-visited denoising states using a teacher with cleaner,
data-assisted temporal context.
{TA-OPD consolidates prediction and refinement specialists into a shared
causal model through local velocity matching under role-specific
sampling and history policies.}

\section{Method}
\label{sec:method}

\begin{figure*}[t]
    \centering
    \includegraphics[width=\linewidth]{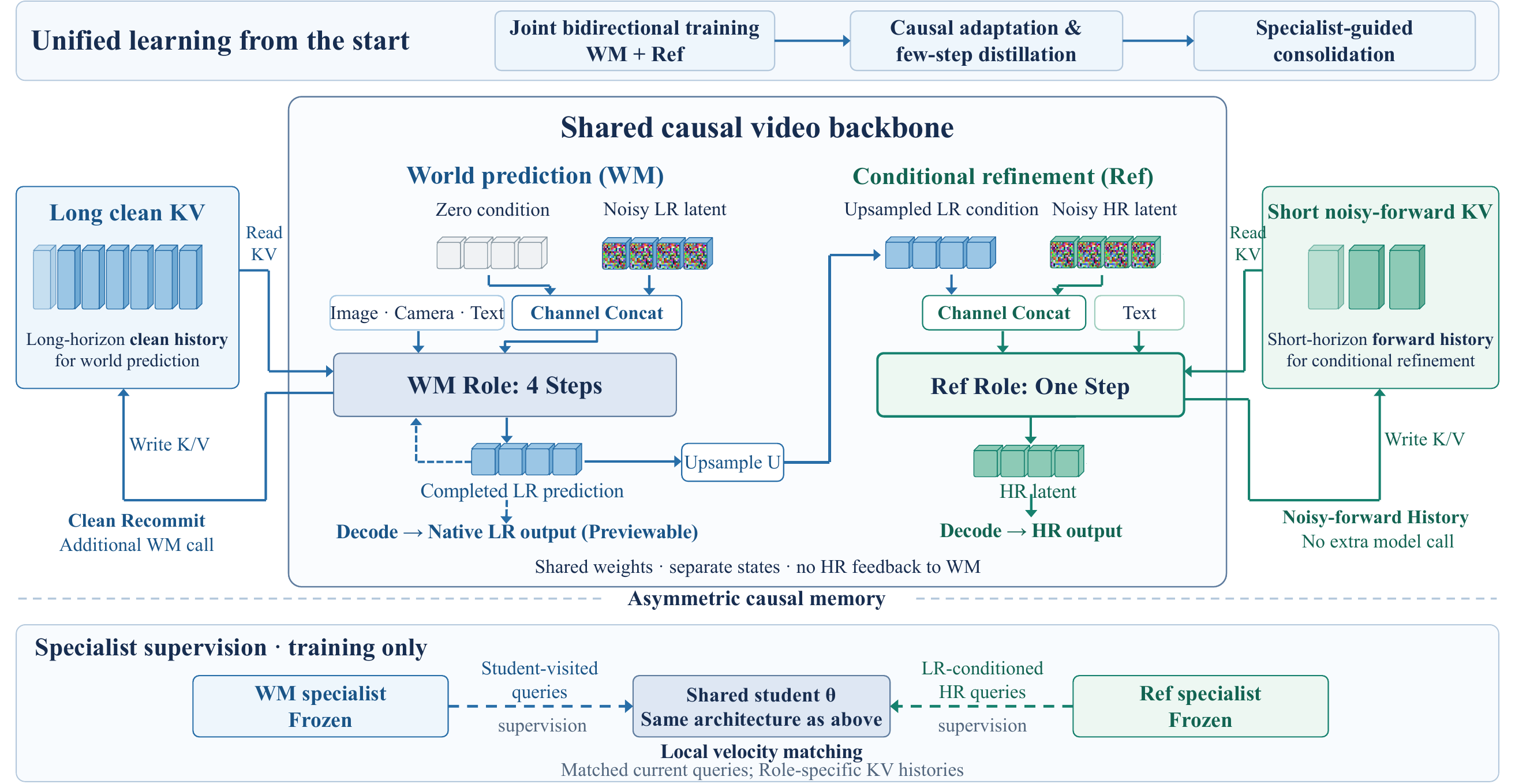}
    \caption{{Overview of \ourmodel{}.}
    \textbf{Top:} Joint prediction and refinement training precedes
    causal adaptation, few-step specialist training, and TA-OPD.
    \textbf{Center:} One shared backbone completes each camera-controlled
    LR prediction and then refines it using the upsampled latent as a
    condition. WM maintains longer clean history; refinement retains
    shorter noisy-forward history. HR outputs do not enter the WM history.
    \textbf{Bottom:} During consolidation, frozen specialists supervise
    the shared student through local velocity matching, with matched
    current queries and network-specific KV histories.}
    \label{fig:pipeline}
\end{figure*}

Figure~\ref{fig:pipeline} summarizes \ourmodel{}.
We first define the shared streaming interface, then specify the two
causal states, explain specialist training and consolidation, and analyze
streaming execution and resource costs.

\subsection{Shared Streaming Prediction and Refinement}
\label{sec:method_overview}
\label{sec:sampling}

We implement world prediction (WM) and high-resolution refinement (Ref)
as two chunk-wise causal samplers with a shared velocity network $v_\theta$.
Our implementation conditions prediction on an initial observation $I$,
text $c$, and camera controls $a$; refinement is learned through video
{super-resolution}.
For chunk $k$, the model first completes a low-resolution (LR) latent
$\widehat z_k^L$ and then generates its high-resolution (HR) counterpart:
\begin{align}
    \widehat z_k^L
    &= S_\theta^{\mathrm{WM}}
    (\epsilon_k^L;I,c,a_{\leq k},M_{k-1}^{\mathrm{WM}}),
    \label{eq:wm_interface}\\
    \widehat z_k^H
    &= S_\theta^{\mathrm{Ref}}
    (\epsilon_k^H;c,U(\widehat z_k^L),M_{k-1}^{\mathrm{Ref}}).
    \label{eq:sr_interface}
\end{align}
Here $U$ spatially upsamples the LR latent, $\epsilon_k^L$ and
$\epsilon_k^H$ collect the samplers' random draws, and
$M_{k-1}^{\mathrm{WM}}$ and $M_{k-1}^{\mathrm{Ref}}$ contain their
separate histories before chunk $k$.
Both roles are causal across chunks: refinement uses the completed
current LR chunk and preceding context, without accessing future LR chunks.
The LR output remains independently usable, and HR outputs never enter
prediction history.

For fixed parameters, supplied camera controls, and independent sampling
streams, the sequence distribution factorizes as
\begin{equation}
\begin{split}
    p_\theta(z^L,z^H\mid I,c,a)
    ={}&p_\theta^{\mathrm{WM}}(z^L\mid I,c,a)\\
       &{}\cdot p_\theta^{\mathrm{Ref}}(z^H\mid z^L,c).
\end{split}
\label{eq:factorization}
\end{equation}
{Internal random variables and recurrent states are marginalized;
refinement operates causally across chunks (Eq.~\ref{eq:sr_interface}).}

\textbf{Shared visual interface.}
Both roles share a causal video backbone and its output head.
The upsampled LR latent is concatenated with the
noisy target before patch embedding. WM sets these condition channels
to zero and uses image anchoring, text, and camera poses injected through
Unified Camera Positional Encoding (UCPE) \citep{zhang2026unified}.
Ref supplies the LR latent and text, with no direct camera input or clean
HR anchor. {The expanded projection retains the original weights and
zero-initializes the new condition weights, preserving the WM function
at initialization (Appendix~\ref{sec:supp_zero_concat}).}

\begin{table}[t]
\centering
\small
\caption{Prediction and refinement share parameters while using distinct conditions, sampling procedures, and causal memories.}
\label{tab:shared_execution}
\setlength{\tabcolsep}{3pt}
\begin{tabular*}{\columnwidth}{@{\extracolsep{\fill}}lcc@{}}
\hline
Property & WM & Ref \\
\hline
Output resolution & low & high \\
Conditions & image, camera, text & LR latent, text \\
Sampling & multi-step & single-step \\
History span & long & short \\
History construction & clean recommit & noisy-forward \\
\hline
\end{tabular*}
\end{table}

\textbf{Role-specific sampling.}
For $m\in\{\mathrm{WM},\mathrm{Ref}\}$, let $c_m$ collect the corresponding
conditions. Under the flow parameterization
$z_t=(1-\sigma_t)z_0+\sigma_t\epsilon$, the network predicts
\begin{equation}
    \widehat z_0 =
    z_t-\sigma_t v_\theta(z_t,t;c_m,M^m).
    \label{eq:clean_prediction}
\end{equation}
WM uses a sequence of denoising queries, re-noising each clean prediction at the
next noise level:
\begin{equation}
    z_{k,j+1}^{\theta,\mathrm{WM}}
    =(1-\sigma_{j+1})\widehat z_{k,j}^{\theta,\mathrm{WM}}
      +\sigma_{j+1}\epsilon_{k,j+1}.
    \label{eq:transition}
\end{equation}
Here $j$ indexes queries, and the last clean prediction becomes the
completed LR chunk. Ref uses one conditional noise query to produce
its HR counterpart. Table~\ref{tab:shared_execution} summarizes the two roles;
their state updates are defined next.

\subsection{Asymmetric Causal Memory}
\label{sec:causal_states}

We design an asymmetric causal memory around the different information
available to the two roles. WM predicts unobserved scene evolution from
preceding context, whereas Ref receives the current scene structure and
motion through its LR condition. We therefore couple a long clean LR
history for prediction with a short noisy-forward HR history for refinement.
The design is asymmetric in both temporal span and history construction.
Window capacities are implementation choices specified in
Appendix~\ref{sec:supp_role_state}.

\textbf{Clean KV for world prediction.}
WM re-encodes each completed LR chunk at zero noise to obtain clean KV
features $K_{k,\mathrm{clean}}^{\mathrm{WM}}$ for subsequent prediction.
With $\mathcal A_m$ denoting the append-and-window operation for role $m$,
\begin{equation}
M_k^{\mathrm{WM}}=\mathcal A_{\mathrm{WM}}
(M_{k-1}^{\mathrm{WM}},K_{k,\mathrm{clean}}^{\mathrm{WM}}).
\label{eq:wm_state_update}
\end{equation}
{Clean recommit adds one LR evaluation after denoising.}

\textbf{Noisy-forward KV reuse for refinement.}
Applying clean recommit to Ref would require a second HR model evaluation
solely to construct history after its one-step generation.
To avoid this cost, we reuse the KV features
$K_{k,\mathrm{fwd}}^{\mathrm{Ref}}$ computed within the refinement forward
that generates the current HR chunk:
\begin{equation}
 M_k^{\mathrm{Ref}}=\mathcal A_{\mathrm{Ref}}
 (M_{k-1}^{\mathrm{Ref}},K_{k,\mathrm{fwd}}^{\mathrm{Ref}}).
 \label{eq:ref_state_update}
\end{equation}
These features are computed from the noisy HR input together with the
current LR condition and preceding refinement context. They are retained
from that forward pass, rather than obtained by re-encoding the completed
HR output $\widehat z_k^H$. Subsequent chunks attend to this memory within
the short refinement window. Thus, each HR denoising evaluation serves
two purposes: generating the current output and providing temporal
features for later chunks. The resulting memory construction removes
the additional HR clean-recommit evaluation while keeping prediction
history and its update rule unchanged.

\textbf{Learning with noisy-forward memory.}
The refinement specialist learns with this recurrent memory.
TA-OPD preserves the same KV construction when consolidating refinement
into the shared student, and streaming inference follows that rule.
This exposes the model during training to the type of history it will
reuse at inference. {Section~\ref{sec:ablation} reports a controlled ablation of
refinement-history construction; Appendix~\ref{sec:supp_kv_comparison}
gives the protocol.}

For window advancement, both roles reindex temporal RoPE to local positions
and transform cached keys without another model evaluation. {Training and inference share this rule;
Appendix~\ref{sec:supp_rope} details RoPE and retained anchors.}

\subsection{Task-Asymmetric On-Policy Distillation}
\label{sec:training}

We introduce \emph{task-asymmetric on-policy distillation} (TA-OPD) to
consolidate separately optimized prediction and refinement behaviors into
one shared model. This separation is motivated by the conflicting gradient
directions and imbalanced gradient magnitudes observed when jointly
optimizing the specialist objectives
(Appendix~\ref{sec:supp_gradient_conflict}). {Building on on-policy
distillation \citep{li2026diffusionopd,zhou2026danceopd}, TA-OPD matches
specialist velocities at the student's current queries under each task's
sampling and memory policies.}

\textbf{Specialist preparation.}
Joint bidirectional training first establishes both task interfaces,
followed by causal adaptation and consistency distillation.
From this shared foundation, we optimize a prediction specialist
$\phi_{\mathrm{WM}}$ and a refinement specialist $\phi_{\mathrm{Ref}}$
through distribution matching distillation (DMD) \citep{yin2024dmd},
using their respective sampling and causal-memory policies.
The refinement specialist additionally uses an auxiliary reconstruction
objective, whereas the subsequent TA-OPD stage uses only local velocity
matching.
Both specialists are then frozen, and the shared student $\theta$ is
initialized from the prediction specialist.

\textbf{Task-asymmetric supervision.}
For WM, the specialist supervises every student-visited denoising query
in each content block.
After each chunk, teacher and student separately encode the same completed
student LR prediction at zero noise to build their own clean KV states.
Supervision follows the student's trajectory and generated history,
with each network constructing its own features.

For Ref, teacher and student process the same conditional HR noise query
using their own recurrent histories and retain their own noisy-forward KV.
Both networks recursively process preceding chunks to construct their
own histories, while only the final block contributes to the
velocity-matching loss. LR conditions during consolidation are obtained from degraded
real videos, and the single-step HR noise input is sampled independently
of the student. Student dependence in refinement therefore enters through
its recurrent feature states. WM additionally follows a student-generated
denoising trajectory.
Within each task, teacher and student match the current noisy latent,
timestep, conditions, window capacity, and RoPE rule, while their KV
states remain network-specific.

\textbf{Local velocity matching.}
For chunk $k$ and query $j$, let $z_{k,j}^{\theta,m}$ be the student's
current noisy latent and $t_j^m$ its timestep. For
$\psi\in\{\theta,\phi_m\}$, define
\begin{equation}
    q_{\psi,k,j}^{m}
    =v_\psi\bigl(
       \operatorname{sg}[z_{k,j}^{\theta,m}],t_j^m;
       c_{m,k},\operatorname{sg}[M_{\psi,k,j}^{m}]
       \bigr).
    \label{eq:opd_query}
\end{equation}
Here $c_{m,k}$ collects the chunk's conditions,
$M_{\psi,k,j}^m$ is network $\psi$'s state before the query, and
$\operatorname{sg}$ stops gradients through earlier sampling and history
construction. Let $d_\theta^m$ denote the joint distribution of queries
and network states generated by the above procedure, with $\tau$ one
realization. The consolidation objective is
\begin{equation}
\begin{split}
    \mathcal L_{\mathrm{con}}
    &=\sum_m p_m\,\mathbb E_{\tau\sim d_\theta^m}
      [\mathcal L_m(\tau)],\\
    \mathcal L_m(\tau)
    &=\frac{1}{|\mathcal B_m|}
      \sum_{k\in\mathcal B_m}\sum_{j=1}^{J_m} w_j^m\,\ell_{k,j}^m,\\
    \ell_{k,j}^m
    &=\operatorname{MSE}
      \bigl(q_{\theta,k,j}^{m},
            \operatorname{sg}[q_{\phi_m,k,j}^{m}]\bigr).
\end{split}
\label{eq:opd}
\end{equation}
Here $\mathcal B_m$ is the set of supervised blocks, $J_m$ is the
number of denoising queries ($J_{\mathrm{Ref}}=1$), and $w_j^m=1/J_m$.
MSE is averaged over latent elements to avoid scaling the loss directly
with spatial resolution. The coefficient $p_m$ denotes the frequency
of task-$m$ updates, rather than an additional per-update loss multiplier.
The sampling schedules and task-update frequencies are specified in
Appendix~\ref{sec:supp_impl}.

Although $d_\theta^m$ changes as the student evolves, each update treats
the sampled queries and cached histories as fixed. Gradients pass only
through the current student velocity prediction, not through the rollout,
state construction, or the parameter dependence of $d_\theta^m$.
TA-OPD adds no reconstruction or perceptual-reward terms.
Rollout lengths, supervised-block selection, and specialist-training
settings are specified in Appendix~\ref{sec:supp_impl}.

\begin{figure*}[!t]
    \centering
    \includegraphics[width=\linewidth]{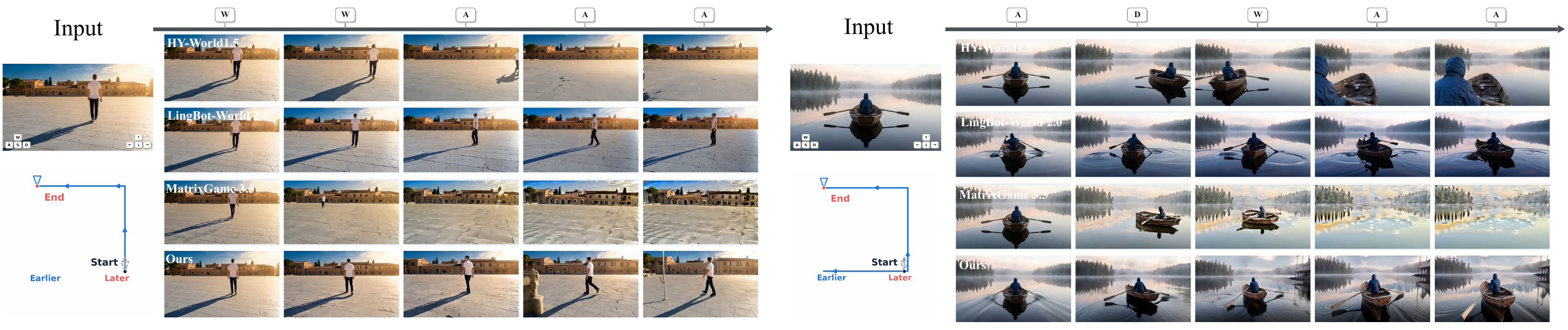}
    \caption{\textbf{Native world prediction.} Initial observations, camera trajectories, and generated frames from HY-World 1.5, LingBot-World 2.0, Matrix-Game 3.5, and our WM role.}
    \label{fig:native_wm_qualitative}
\end{figure*}

\begin{figure*}[!t]
    \centering
    \includegraphics[width=\linewidth]{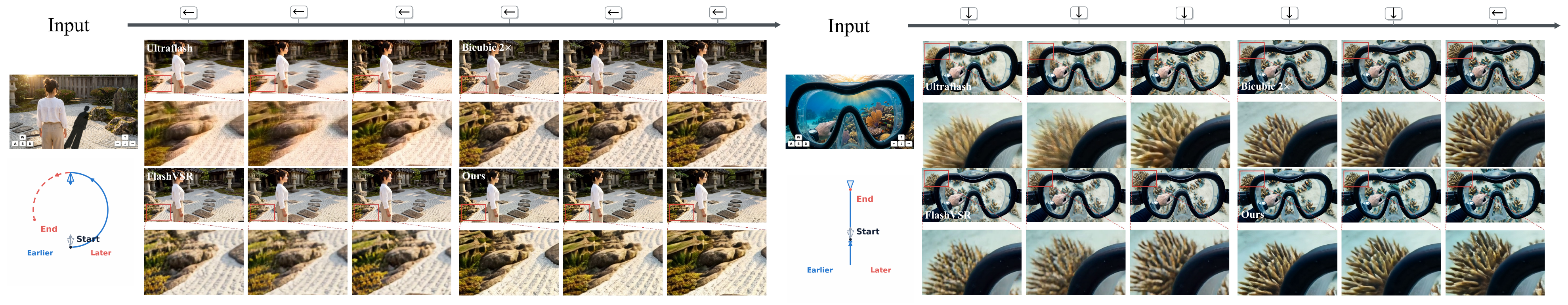}
    \caption{\textbf{High-resolution refinement.} Ultra Flash, bicubic $2\times$, FlashVSR, and our model on two examples; enlarged regions expose differences in structure and texture.}
    \label{fig:sr_qualitative}
\end{figure*}

\subsection{Streaming Inference}
\label{sec:inference}

At inference, the shared model alternates between prediction and
refinement. Each completed LR chunk updates the clean prediction memory
and conditions its HR counterpart; the refinement forward simultaneously
produces the HR output and the noisy-forward KV for subsequent chunks.
This process uses only the available LR prefix and follows the same
causal-memory rules as training. With fixed parameters, independent
sampling streams, and unchanged prediction settings, enabling refinement
leaves the native LR sample path unchanged.

\textbf{Memory and computation.}
The asymmetric memory allocates long-term context at low resolution
while restricting the more expensive HR history to a shorter window.
Let $n_L,n_H$ denote spatial tokens per latent frame and $w,s$ the
prediction and refinement window capacities, including the current chunk
and any retained anchor. For the same backbone, cache precision, and
prediction-window capacity $w$, the self-attention KV-payload ratio
relative to direct-HR prediction is
\begin{equation}
    \frac{M_{\mathrm{KV}}^{\mathrm{cascade}}}
         {M_{\mathrm{KV}}^{\mathrm{direct\text{-}HR}}}
    = \frac{n_Lw+n_Hs}{n_Hw}
    = \frac{n_L}{n_H}+\frac{s}{w}.
    \label{eq:kv_ratio}
\end{equation}
With $s$ fixed, each additional prediction-window slot costs only
$n_L/n_H$ as much KV storage as a direct-HR slot. {This ratio covers only KV storage; total GPU memory also includes
weights and other allocations (Appendix~\ref{sec:supp_kv_budget}).}

Parameter sharing removes the duplicate backbone weights required by
separate prediction and refinement specialists, while both causal
memories are retained. {Noisy-forward KV reuse additionally saves computation by removing the
extra HR clean-recommit forward pass.} 

\FloatBarrier
\section{Experiments}
\label{sec:experiments}

\subsection{Experimental Setup}
\label{sec:exp_setup}
{We build \ourmodel{} on Wan2.2 TI2V-5B, with world
prediction at $480{\times}832$ and streaming refinement at
$960{\times}1664$.}

We evaluate native world prediction and final
high-resolution generation on WBench~\citep{ying2026wbench}, reporting
overall and per-dimension scores. All methods receive the same
observations, camera commands, and temporal intervals, while retaining
their supported inference settings. Native prediction baselines include
HY-World 1.5~\citep{sun2025worldplay},
LingBot-World 2.0 (fast)~\citep{gao2026infiniteworlds}, and
Matrix-Game 3.5~\citep{qian2026matrixgame35}. For high-resolution
generation, we compare against WorldPlay~\citep{sun2025worldplay},
LingBot-World 2.0~\citep{gao2026infiniteworlds}, and
Matrix-Game 3.5~\citep{qian2026matrixgame35} paired with
Ultra Flash~\citep{luxury2026ultraflash} or
FlashVSR~\citep{zhuang2025flashvsr}.

We evaluate refinement separately using the same fixed
set of decoded LR videos generated by our world model. We report
MUSIQ \citep{ke2021musiq} and CLIP-IQA+ \citep{wang2023clipiqa} for
perceptual quality, LR-consistency PSNR for preservation of the input
content, and motion-compensated residuals (MC L1 and MC HF L1) for
temporal variation. {Appendices~\ref{sec:supp_impl} and~\ref{sec:supp_eval}
detail training, evaluation protocols, and metrics.}

\subsection{Generation Quality and Capability Retention}
\label{sec:main_results}
\textbf{Shared versus separate generation.}
{Table~\ref{tab:hr_system}(a) compares the unified model
with a cascade of separately optimized specialists. The noisy-forward shared model
achieves comparable overall HR quality, with stronger setting and
interaction adherence and slightly lower visual quality. These results
suggest that prediction and refinement can coexist in one backbone;
the matched deployment comparison below isolates the resource savings
from sharing its weights.}

\begin{table*}[t]
\centering
\small
\caption{{WBench generation quality and capability retention.}
(a) Complete HR generation with $480{\times}832$ prediction and $960{\times}1664$ refinement.
(b) Native world prediction without refinement.
{$W$, $R$, and $U$ denote the WM specialist, refinement specialist, and unified model.}
Bold marks the highest score per metric within each group.}
\label{tab:hr_system}
\label{tab:native_wm}
\setlength{\tabcolsep}{3pt}
\renewcommand{\arraystretch}{1.0}
\begin{tabularx}{\textwidth}{@{}l*{6}{>{\centering\arraybackslash}X}@{}}
\toprule
Method & Average$\uparrow$ & \shortstack{Video\\Quality$\uparrow$}
& \shortstack{Setting\\Adherence$\uparrow$}
& \shortstack{Interaction\\Adherence$\uparrow$}
& Consistency$\uparrow$ & Physical$\uparrow$ \\
\midrule
\multicolumn{7}{@{}l}{\textbf{(a) Complete HR generation}} \\
Separate cascade ($W\rightarrow R$) & {81.58} & {83.72} & {78.21} & 86.80 & \textbf{{87.04}} & \textbf{{72.13}} \\
Shared $U\rightarrow U$ (clean-recommit KV) & {81.53} & \textbf{{83.83}} & {77.81} & 86.98 & {87.01} & {72.03} \\
Shared $U\rightarrow U$ (noisy-forward KV) & \textbf{{81.60}} & 83.43 & \textbf{{78.73}} & \textbf{87.08} & {86.92} & {71.82} \\
\midrule
\multicolumn{7}{@{}l}{\textbf{(b) Native world prediction without refinement}} \\
Matrix-Game 3.5 & 74.760 & 78.46 & 72.10 & 83.12 & 79.63 & 60.51 \\
HY-World 1.5 & 79.150 & 79.82 & 72.79 & \textbf{87.38} & \textbf{87.52} & 68.25 \\
LingBot-World 2.0 (fast) & \textbf{81.460} & 82.23 & \textbf{80.50} & 84.92 & 84.45 & \textbf{75.22} \\
WM-only specialist $W$ & 80.522 & 83.02 & 77.14 & 85.91 & 86.52 & 70.02 \\
Unified WM $U$ & 80.308 & \textbf{83.04} & 77.28 & 86.05 & 85.64 & 69.53 \\
\bottomrule
\end{tabularx}
\end{table*}

\textbf{{Capability retention.}}
{The unified model remains close to its WM specialist
on native WBench performance and achieves the highest visual-quality
score among the evaluated native models (Table~\ref{tab:native_wm}(b)).
The modest decrease in consistency shows that retention is not uniform
across dimensions. For refinement, the shared noisy-forward model
remains close to its independent specialist on MUSIQ under the same
decoded-LR protocol (Table~\ref{tab:supp_hr_system}).
Figure~\ref{fig:native_wm_qualitative} compares native
camera-controlled outputs across models.}

\textbf{Complete HR systems.}
{Our unified prediction--refinement pipeline achieves the highest
overall WBench score (81.60) among all evaluated systems, including
native world models and cascaded high-resolution baselines
(Tables~\ref{tab:hr_system} and~\ref{tab:refined_wm}). These scores reflect both prediction and refinement.
We therefore compare refiners on fixed LR inputs below
to control for upstream differences.}

\begin{table*}[!htbp]
\centering
\small
\setlength{\tabcolsep}{4pt}
\caption{WBench scores on final HR videos. Each external WM supplies the same frozen predictions to its two refiners. Bold marks the highest score per column.}
\label{tab:refined_wm}
\begin{tabular*}{\textwidth}{@{\extracolsep{\fill}}llcccccc@{}}
\hline
World model & Refiner & Avg.$\uparrow$ & Qual.$\uparrow$
& Set.$\uparrow$ & Inter.$\uparrow$ & Cons.$\uparrow$ & Phys.$\uparrow$ \\
\hline
WorldPlay & Ultra Flash &
78.630 & 79.69 & 72.27 & 85.62 & \textbf{87.43} & 68.13 \\
WorldPlay & FlashVSR &
79.160 & 80.08 & 73.09 & 86.52 & 87.09 & 69.01 \\
LingBot-World 2.0 & Ultra Flash &
{81.070} & 81.94 & 79.29 & 84.73 & 84.36 & \textbf{75.03} \\
LingBot-World 2.0 & FlashVSR &
81.030 & 81.98 & \textbf{79.86} & 84.93 & 84.44 & 73.96 \\
Matrix-Game 3.5 & Ultra Flash &
73.990 & 78.56 & 70.81 & 83.43 & 79.48 & 57.69 \\
Matrix-Game 3.5 & FlashVSR &
75.060 & 78.72 & 72.53 & 83.14 & 79.51 & 61.41 \\
\hline
\ourmodel{} WM (ours) & Shared Ref (noisy-forward KV) &
\textbf{81.596} & \textbf{83.43} & 78.73 & \textbf{87.08} & 86.92 & 71.82 \\
\hline
\end{tabular*}
\end{table*}

\begin{table}[!tb]
\centering
\small
\setlength{\tabcolsep}{2pt}
\caption{{Refinement of identical frozen LR videos from our WM. Compare rows only within each spatial protocol. Bold marks the best full-frame scores. PSNR measures LR consistency, not HR reconstruction; MC residuals use grayscale in $[0,1]$. Clean/noisy denote clean-recommit/noisy-forward KV histories.}}
\label{tab:fixed_lr}
\begin{tabular*}{\columnwidth}{@{\extracolsep{\fill}}lccccc@{}}
\hline
Refiner & MUSIQ$\uparrow$ & \shortstack{CLIP-\\IQA+$\uparrow$}
& \shortstack{LR\\PSNR$\uparrow$} & \shortstack{MC\\L1$\downarrow$}
& \shortstack{MC HF\\L1$\downarrow$} \\
\hline
\multicolumn{6}{@{}l}{FlashVSR center-crop protocol: $896{\times}1664$ output} \\
FlashVSR Full & 61.9861 & {0.5167} & 23.4840 & 0.0327 & 0.0140 \\
\hline
\multicolumn{6}{@{}l}{Full frame: $480{\times}832 \rightarrow 960{\times}1664$} \\
Ultra Flash v1.1 & 53.7366 & 0.4417 & 22.2844 & 0.0330 & 0.0090 \\
{Ours (clean)} & 56.7657 & 0.5156 & 25.3340 & {\textbf{0.0208}} & {\textbf{0.0081}} \\
{Ours (noisy)} & \textbf{58.9656} & \textbf{0.5185} & {\textbf{26.2409}} & 0.0212 & 0.0086 \\
\hline
\end{tabular*}
\end{table}

\textbf{Refinement of fixed predictions.}
{On the same generated LR inputs, our refiner improves
perceptual quality, LR consistency, and motion-compensated residuals
over Ultra Flash under the full-frame protocol (Table~\ref{tab:fixed_lr}).
FlashVSR's elevated MC L1 and MC HF L1 residuals suggest temporal
fluctuations consistent with inter-frame flicker, although its
center-crop protocol limits direct numerical comparison. Our outputs
exhibit smoother frame-to-frame transitions and greater temporal
continuity. Figure~\ref{fig:sr_qualitative} shows qualitative
comparisons.}

\begin{figure*}[!t]
    \centering
    \includegraphics[width=\textwidth]{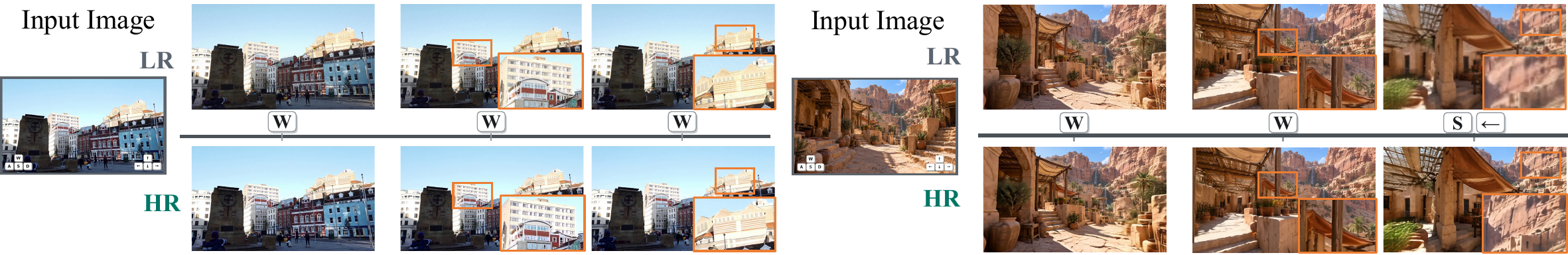}
    \caption{{\textbf{Qualitative self-cascade results.}
    Paired LR/HR outputs with camera controls; orange insets show matched regions.}}
    \label{fig:main_more_results}
\end{figure*}

{Figure~\ref{fig:main_more_results} illustrates our unified
prediction--refinement pipeline on camera-controlled rollouts. The HR
outputs preserve the scene layout of the LR predictions while
producing sharper structural boundaries and richer local textures,
as highlighted by the matched crops.}

\begin{table}[!tb]
\centering
\small
\setlength{\tabcolsep}{4pt}
\caption{{Ablation of distillation strategies for native WM prediction. All variants use identical initialization.}}
\label{tab:training}
\begin{tabular*}{\columnwidth}{@{\extracolsep{\fill}}lccc@{}}
\hline
Metric & \shortstack{Mixed\\DMD} & \shortstack{CD +\\mixed DMD} & TA-OPD \\
\hline
Avg.$\uparrow$ & 72.000 & 75.878 & \textbf{{80.308}} \\
Qual.$\uparrow$ & 79.82 & 75.72 & \textbf{83.04} \\
Set.$\uparrow$ & 63.94 & 64.53 & \textbf{77.28} \\
Inter.$\uparrow$ & 76.79 & \textbf{87.98} & 86.05 \\
Cons.$\uparrow$ & 79.96 & 82.44 & \textbf{85.64} \\
Phys.$\uparrow$ & 59.49 & 68.72 & \textbf{69.53} \\
\hline
\end{tabular*}
\end{table}

\subsection{Training and Efficiency Analysis}
\label{sec:ablation}
\textbf{{Distillation strategies.}}
{We compare mixed DMD, CD + mixed DMD, and TA-OPD
from identical initialization (Table~\ref{tab:training}).
TA-OPD achieves the highest overall native prediction score and
setting adherence. In contrast, mixed DMD and CD + mixed DMD
suffer from cross-task gradient interference during joint distillation
(Table~\ref{tab:mixed_dmd_gradients}), limiting their overall performance.}

\begin{table}[!htbp]
\centering
\small
\caption{WM/{Ref} DMD gradient alignment during mixed training. {Negative blocks have cosine $<0$; probe settings: Appendix~\ref{sec:supp_gradient_conflict}.}}
\label{tab:mixed_dmd_gradients}
\begin{tabular*}{\columnwidth}{@{\extracolsep{\fill}}rcc@{}}
\hline
Training step & Global cosine & Negative blocks \\
\hline
400 & $+0.0969$ & 2/30 \\
600 & $-0.1748$ & 29/30 \\
700 & $-0.1775$ & 25/30 \\
800 & $-0.3421$ & 30/30 \\
\hline
\end{tabular*}
\end{table}

\textbf{Task interference in mixed distillation.}
{WM and {Ref} gradients shift from weak global alignment
to opposing directions across the backbone during mixed distillation
(Table~\ref{tab:mixed_dmd_gradients}). These measurements show conflicting
update directions in the sampled training states and motivate separate
specialist optimization before consolidation.
Appendix~\ref{sec:supp_gradient_conflict} details the diagnostic
protocol and output analysis.}

\begin{table}[!htbp]
\centering
\small
\caption{{Controlled ablation of refinement-history construction on fixed LR inputs.} HR forwards include history construction; {generator time is measured per 16-frame chunk before VAE decoding; generator FPS is 16 divided by this time.} PSNR measures LR consistency; HR Q is MUSIQ and TE is MC L1.
{All other training and evaluation settings are identical.}}
\label{tab:kv_config}
\setlength{\tabcolsep}{4pt}
\begin{tabular*}{\columnwidth}{@{\extracolsep{\fill}}lcc@{}}
\hline
Metric & \shortstack{Clean-\\recommit} & \shortstack{Noisy-forward\\(reference)} \\
\hline
HR forwards & 2 & 1 \\
HR Q$\uparrow$ & 56.7657 & \textbf{58.9656} \\
PSNR$\uparrow$ & 25.3340 & \textbf{26.2409} \\
TE$\downarrow$ & \textbf{0.0208} & 0.0212 \\
{Generator time (s)$\downarrow$} & {0.60} & {\textbf{0.30}} \\
{Generator FPS$\uparrow$} & {26.7} & {\textbf{53.3}} \\
\hline
\end{tabular*}
\end{table}

\textbf{Noisy-forward refinement history.}
{Table~\ref{tab:kv_config} compares clean-recommit and
noisy-forward histories with all other training and evaluation settings
fixed. Reusing KV from the generation forward eliminates the additional
HR pass used for clean recommit, yielding a $2\times$ generator speedup
before VAE decoding. End-to-end throughput improves by 26.9\%, including
WM prediction, refinement, history construction, and decoding
(Table~\ref{tab:supp_hr_system}). Noisy-forward history also improves
MUSIQ and LR consistency on fixed inputs, with a small increase in MC L1.
It thus reduces computation while improving perceptual quality, with a
measured trade-off in temporal residuals.
Appendix~\ref{sec:supp_kv_comparison} gives the full settings.}

\begin{table}[H]
\centering
\small
\setlength{\tabcolsep}{3pt}
\caption{Measured deployment costs on four GPUs (BF16, batch size one). Weight/KV are per-GPU allocations in GiB; Peak is the maximum allocated memory across GPUs. FPS covers end-to-end generation, including decoding. }
\label{tab:deployment}
\begin{tabular*}{\columnwidth}{@{\extracolsep{\fill}}lccc@{}}
\hline
Metric & \shortstack{$W\rightarrow R$\\Resident} & \shortstack{$U\rightarrow U$\\Two copies} & \shortstack{$U\rightarrow U$\\Shared copy} \\
\hline
Copies & 2 & 2 & 1 \\
Params (B) & 10.28 & 10.28 & 5.14 \\
Weight GiB & 19.16 & 19.16 & 9.58 \\
KV GiB & 1.77 & 1.77 & 1.77 \\
Peak GiB$\downarrow$ & 45.42 & 45.42 & \textbf{35.84} \\
FPS$\uparrow$ & 10.77 & \textbf{11.94} & \textbf{11.94} \\
\hline
\end{tabular*}
\end{table}

\textbf{Shared-backbone deployment.}
{Table~\ref{tab:deployment} compares one- and two-copy deployments
of the same unified checkpoint under matched inference settings
(Appendix~\ref{sec:efficiency}). Weight sharing halves resident backbone
storage and reduces peak GPU memory by 21.1\% while preserving
end-to-end throughput. KV allocation remains unchanged because both
role-specific causal states are retained; the memory savings therefore
come from removing duplicate weights.}

{Parameter sharing and asymmetric memory address
different costs: the former reduces resident weights, while the latter
stores the longer prediction history at low resolution. The direct-HR
comparison in Eq.~\ref{eq:kv_ratio} concerns KV payload and is distinct
from the measured deployment savings.}

\section{Conclusion}
\label{sec:conclusion}
We presented \ourmodel{}, a shared causal model for camera-controlled
prediction and high-resolution refinement. Its asymmetric temporal
contexts allocate long history to low-resolution prediction, while
specialist training and TA-OPD integrate the two capabilities.
Experiments show comparable overall generation quality to a separate
specialist cascade, retained native prediction capability, and lower
resident deployment memory. The resulting model produces high-resolution worlds
incrementally with one backbone. Limitations are discussed in
Appendix~\ref{sec:limitations}.

{\small
\bibliography{main}

@String(PAMI = {IEEE Trans. Pattern Anal. Mach. Intell.})

@String(CVPR = {IEEE Conf. Comput. Vis. Pattern Recog.})

@String(ICCV = {Int. Conf. Comput. Vis.})

@String(NIPS = {Adv. Neural Inform. Process. Syst.})

@String(ICLR = {Int. Conf. Learn. Represent.})

@String(AAAI = {AAAI})

@inproceedings{huang2025selfforcing,
  title   = {Self Forcing: Bridging the Train-Test Gap in Autoregressive Video Diffusion},
  author  = {Huang, Xun and Li, Zhengqi and He, Guande and Zhou, Mingyuan and Shechtman, Eli},
  booktitle = NIPS,
  year    = {2025},
  url     = {https://proceedings.neurips.cc/paper_files/paper/2025/hash/f4823f831af67a3ef15e41a85434422a-Abstract-Conference.html}
}

@article{sun2025worldplay,
  title   = {WorldPlay: Towards Long-Term Geometric Consistency for Real-Time Interactive World Modeling},
  author  = {Sun, Wenqiang and Zhang, Haiyu and Wang, Haoyuan and Wu, Junta and Wang, Zehan and Wang, Zhenwei and Wang, Yunhong and Zhang, Jun and Wang, Tengfei and Guo, Chunchao},
  journal = {arXiv preprint arXiv:2512.14614},
  year    = {2025},
  url     = {https://arxiv.org/abs/2512.14614}
}

@article{gao2026infiniteworlds,
  title   = {Infinite Worlds with Versatile Interactions},
  author  = {Gao, Zelin and Wang, Qiuyu and Zhu, Jiapeng and Chen, Jingye and Liu, Zichen and Bai, Qingyan and Wang, Jiahao and Yuan, Yufeng and Wang, Hanlin and Lu, Yichong and others},
  journal = {arXiv preprint arXiv:2607.07534},
  year    = {2026}
}

@article{luxury2026ultraflash,
  title   = {Ultra Flash: Scaling Real-Time Streaming Video Generation to High Resolutions},
  author  = {Huang, Jie and Lu, Xin and Fan, Zihao and Ma, Xiaoxiao and Zhuang, Jun-hao and Li, Yuming and Xue, Zeyue and Fu, Siming and Li, Haoran and Zhong, Mingchen and others},
  journal = {arXiv preprint arXiv:2606.09150},
  year    = {2026}
}

@inproceedings{ho2022videodiffusion,
  title     = {Video Diffusion Models},
  author    = {Ho, Jonathan and Salimans, Tim and Gritsenko, Alexey and Chan, William and Norouzi, Mohammad and Fleet, David J.},
  booktitle = NIPS,
  year      = {2022}
}

@inproceedings{yin2025causvid,
  title     = {From Slow Bidirectional to Fast Autoregressive Video Diffusion Models},
  author    = {Yin, Tianwei and Zhang, Qiang and Zhang, Richard and Freeman, William T. and Durand, Fr{\'e}do and Shechtman, Eli and Huang, Xun},
  booktitle = CVPR,
  year      = {2025}
}

@inproceedings{zhu2026causalforcing,
  title     = {Causal Forcing: Autoregressive Diffusion Distillation Done Right for High-Quality Real-Time Interactive Video Generation},
  author    = {Zhu, Hongzhou and Zhao, Min and He, Guande and Su, Hang and Li, Chongxuan and Zhu, Jun},
  booktitle = {International Conference on Machine Learning},
  year      = {2026}
}

@article{wan2025,
  title   = {Wan: Open and Advanced Large-Scale Video Generative Models},
  author  = {{Team Wan} and Wang, Ang and Ai, Baole and Wen, Bin and Mao, Chaojie and Xie, Chen-Wei and Chen, Di and Yu, Feiwu and Zhao, Haiming and Yang, Jianxiao and others},
  journal = {arXiv preprint arXiv:2503.20314},
  year    = {2025}
}

@article{he2024venhancer,
  title   = {{VEnhancer}: Generative Space-Time Enhancement for Video Generation},
  author  = {He, Jingwen and Xue, Tianfan and Liu, Dongyang and Lin, Xinqi and Gao, Peng and Lin, Dahua and Qiao, Yu and Ouyang, Wanli and Liu, Ziwei},
  journal = {arXiv preprint arXiv:2407.07667},
  year    = {2024}
}

@article{zhuang2025flashvsr,
  title   = {{FlashVSR}: Towards Real-Time Diffusion-Based Streaming Video Super-Resolution},
  author  = {Zhuang, Junhao and Guo, Shi and Cai, Xin and Li, Xiaohui and Liu, Yihao and Yuan, Chun and Xue, Tianfan},
  journal = {arXiv preprint arXiv:2510.12747},
  year    = {2025}
}

@inproceedings{zhang2026unified,
  title     = {Unified Camera Positional Encoding for Controlled Video Generation},
  author    = {Zhang, Cheng and Li, Boying and Wei, Meng and Cao, Yan-Pei and Gambardella, Camilo and Phung, Dinh and Cai, Jianfei},
  booktitle = {Proceedings of the IEEE/CVF Conference on Computer Vision and Pattern Recognition},
  pages     = {38027--38037},
  year      = {2026}
}

@article{guo2024selfcascade,
  title   = {Make a Cheap Scaling: A Self-Cascade Diffusion Model for Higher-Resolution Adaptation},
  author  = {Guo, Lanqing and He, Yingqing and Chen, Haoxin and Xia, Menghan and Cun, Xiaodong and Wang, Yufei and Huang, Siyu and Zhang, Yong and Wang, Xintao and Chen, Qifeng and Shan, Ying and Wen, Bihan},
  journal = {arXiv preprint arXiv:2402.10491},
  year    = {2024}
}

@inproceedings{yin2024dmd,
  title     = {One-Step Diffusion with Distribution Matching Distillation},
  author    = {Yin, Tianwei and Gharbi, Micha{\"e}l and Zhang, Richard and Shechtman, Eli and Durand, Fr{\'e}do and Freeman, William T. and Park, Taesung},
  booktitle = CVPR,
  year      = {2024}
}

@inproceedings{wang2023clipiqa,
  title     = {Exploring {CLIP} for Assessing the Look and Feel of Images},
  author    = {Wang, Jianyi and Chan, Kelvin C. K. and Loy, Chen Change},
  booktitle = AAAI,
  year      = {2023}
}

@inproceedings{ke2021musiq,
  title     = {{MUSIQ}: Multi-Scale Image Quality Transformer},
  author    = {Ke, Junjie and Wang, Qifei and Wang, Yilin and Milanfar, Peyman and Yang, Feng},
  booktitle = ICCV,
  pages     = {5148--5157},
  year      = {2021}
}

@article{ding2022dists,
  title   = {Image Quality Assessment: Unifying Structure and Texture Similarity},
  author  = {Ding, Keyan and Ma, Kede and Wang, Shiqi and Simoncelli, Eero P.},
  journal = PAMI,
  volume  = {44},
  number  = {5},
  pages   = {2567--2581},
  year    = {2022}
}

@inproceedings{zhang2018lpips,
  title     = {The Unreasonable Effectiveness of Deep Features as a Perceptual Metric},
  author    = {Zhang, Richard and Isola, Phillip and Efros, Alexei A. and Shechtman, Eli and Wang, Oliver},
  booktitle = CVPR,
  pages     = {586--595},
  year      = {2018}
}

@inproceedings{jin2024pyramidflow,
  title   = {Pyramidal Flow Matching for Efficient Video Generative Modeling},
  author  = {Jin, Yang and Sun, Zhicheng and Li, Ningyuan and Xu, Kun and Jiang, Hao and Zhuang, Nan and Huang, Quzhe and Song, Yang and Mu, Yadong and Lin, Zhouchen},
  booktitle = ICLR,
  year    = {2025},
  url     = {https://proceedings.iclr.cc/paper_files/paper/2025/hash/3ab228c4703c4459b1a600ebadc5732c-Abstract-Conference.html}
}

@article{korzhenkov2026pyramidalwan,
  title   = {{PyramidalWan}: On Making Pretrained Video Model Pyramidal for Efficient Inference},
  author  = {Korzhenkov, Denis and Karjauv, Adil and Karnewar, Animesh and Ghafoorian, Mohsen and Habibian, Amirhossein},
  journal = {arXiv preprint arXiv:2601.04792},
  year    = {2026}
}

@article{meituan2025longcat,
  title   = {{LongCat-Video} Technical Report},
  author  = {{Meituan LongCat Team}},
  journal = {arXiv preprint arXiv:2510.22200},
  year    = {2025}
}

@article{qiu2025histream,
  title   = {{HiStream}: Efficient High-Resolution Video Generation via Redundancy-Eliminated Streaming},
  author  = {Qiu, Haonan and Liu, Shikun and Zhou, Zijian and An, Zhaochong and Ren, Weiming and Liu, Zhiheng and Schult, Jonas and He, Sen and Chen, Shoufa and Cong, Yuren and Xiang, Tao and Liu, Ziwei and Perez-Rua, Juan-Manuel},
  journal = {arXiv preprint arXiv:2512.21338},
  year    = {2025}
}

@inproceedings{hur2025hifi,
  title     = {High-Resolution Frame Interpolation with Patch-Based Cascaded Diffusion},
  author    = {Hur, Junhwa and Herrmann, Charles and Saxena, Saurabh and Kontkanen, Janne and Lai, Wei-Sheng and Shih, Yichang and Rubinstein, Michael and Fleet, David J. and Sun, Deqing},
  booktitle = AAAI,
  year      = {2025}
}

@article{wang2025seedvr,
  title   = {{SeedVR}: Seeding Infinity in Diffusion Transformer Towards Generic Video Restoration},
  author  = {Wang, Jianyi and Lin, Zhijie and Wei, Meng and Zhao, Yang and Yang, Ceyuan and Xiao, Fei and Loy, Chen Change and Jiang, Lu},
  journal = {arXiv preprint arXiv:2501.01320},
  year    = {2025}
}

@inproceedings{wang2026seedvr2,
  title   = {{SeedVR2}: One-Step Video Restoration via Diffusion Adversarial Post-Training},
  author  = {Wang, Jianyi and Lin, Shanchuan and Lin, Zhijie and Ren, Yuxi and Wei, Meng and Yue, Zongsheng and Zhou, Shangchen and Chen, Hao and Zhao, Yang and Yang, Ceyuan and Xiao, Xuefeng and Loy, Chen Change and Jiang, Lu},
  booktitle = ICLR,
  year    = {2026},
  url     = {https://proceedings.iclr.cc/paper_files/paper/2026/hash/444d69470b24ded080183c907b711bbf-Abstract-Conference.html}
}

@article{zhu2026sanawm,
  title   = {{SANA-WM}: Efficient Minute-Scale World Modeling with Hybrid Linear Diffusion Transformer},
  author  = {Zhu, Haoyi and Liu, Haozhe and Zhao, Yuyang and Ye, Tian and Chen, Junsong and Yu, Jincheng and He, Tong and Han, Song and Xie, Enze},
  journal = {arXiv preprint arXiv:2605.15178},
  year    = {2026}
}

@article{li2026diffusionopd,
  title   = {{DiffusionOPD}: A Unified Perspective of On-Policy Distillation in Diffusion Models},
  author  = {Li, Quanhao and Yu, Junqiu and Jiang, Kaixun and Wei, Yujie and Xing, Zhen and Li, Pandeng and Chu, Ruihang and Zhang, Shiwei and Liu, Yu and Wu, Zuxuan},
  journal = {arXiv preprint arXiv:2605.15055},
  year    = {2026}
}

@article{zhou2026danceopd,
  title   = {{DanceOPD}: On-Policy Generative Field Distillation},
  author  = {Zhou, Wei and Zhu, Xiongwei and Xu, Zelin and Dong, Bo and Gong, Lixue and Liang, Yongyuan and Chu, Meng and Qu, Leigang and Kong, Lingdong and Liu, Wei and Chua, Tat-Seng},
  journal = {arXiv preprint arXiv:2606.27377},
  year    = {2026}
}

@article{liu2026opsdv,
  title   = {{OPSD-V}: On-Policy Self-Distillation for Post-Training Few-Step Autoregressive Video Generators},
  author  = {Liu, Hongyu and Wang, Chun and Gao, Feng and He, Xuanhua and Ma, Yue and Wan, Ziyu and Zhang, Yong and Wei, Xiaoming and Chen, Qifeng},
  journal = {arXiv preprint arXiv:2607.08766},
  year    = {2026},
  url     = {https://arxiv.org/abs/2607.08766}
}

@article{qian2026matrixgame35,
  title   = {{Matrix-Game 3.5}: Enhancing Real-Time Streaming Interactive World Models with Patch Memory},
  author  = {Qian, Runjia and Wang, Zile and Zhang, Jihai and Zou, Kai and Yu, Wei and Li, Jiaxing and Liu, Zexiang and Li, Yaokun and Kang, Fei and Huang, Kaichen and An, Mengyin and Zhang, Haobo and Jiang, Biao and Wang, Jiahua and Sun, Haofeng and Liu, Yang and Li, Yangguang},
  journal = {arXiv preprint arXiv:2608.29910},
  year    = {2026}
}

@article{ying2026wbench,
  title   = {{WBench}: A Comprehensive Multi-turn Benchmark for Interactive Video World Model Evaluation},
  author  = {Ying, Kaining and Hu, Hengrui and Ren, Siyu and Li, Jiamu and Chen, Fengjiao and Wang, Ziwen and Cao, Xuezhi and Cai, Xunliang and Ding, Henghui},
  journal = {arXiv preprint arXiv:2605.25874},
  year    = {2026}
}

@inproceedings{bruce2024genie,
  title     = {{Genie}: Generative Interactive Environments},
  author    = {Bruce, Jake and Dennis, Michael D and Edwards, Ashley and Parker-Holder, Jack and Shi, Yuge and Hughes, Edward and Lai, Matthew and Mavalankar, Aditi and Steigerwald, Richie and Apps, Chris and others},
  booktitle = {International Conference on Machine Learning},
  year      = {2024},
  url       = {https://proceedings.mlr.press/v235/bruce24a.html}
}

@inproceedings{valevski2025gamengen,
  title     = {Diffusion Models Are Real-Time Game Engines},
  author    = {Valevski, Dani and Leviathan, Yaniv and Arar, Moab and Fruchter, Shlomi},
  booktitle = ICLR,
  year      = {2025},
  url       = {https://arxiv.org/abs/2408.14837}
}

@article{hu2023gaia1,
  title   = {{GAIA-1}: A Generative World Model for Autonomous Driving},
  author  = {Hu, Anthony and Russell, Lloyd and Yeo, Hudson and Murez, Zak and Fedoseev, George and Kendall, Alex and Shotton, Jamie and Corrado, Gianluca},
  journal = {arXiv preprint arXiv:2309.17080},
  year    = {2023},
  url     = {https://arxiv.org/abs/2309.17080}
}

@article{nvidia2025cosmos,
  title   = {{Cosmos} World Foundation Model Platform for Physical {AI}},
  author  = {{NVIDIA} and Agarwal, Niket and Ali, Arslan and Bala, Maciej and Balaji, Yogesh and Barker, Erik and Cai, Tiffany and Chattopadhyay, Prithvijit and Chen, Yongxin and Cui, Yin and Ding, Yifan and others},
  journal = {arXiv preprint arXiv:2501.03575},
  year    = {2025},
  url     = {https://arxiv.org/abs/2501.03575}
}

@article{zhu2026dreamxcreator,
  title   = {{DreamX-Creator}: Democratizing Native Audio-Video Generation at {2K} Resolution},
  author  = {Zhu, Jiashu and Zheng, Yanhao and Tian, Ruitian and Dang, Rujing and Zhang, Shen and Song, Bingze and Lei, Jiachen and Lin, Ruimin and Wu, Jiahong and Chu, Xiangxiang},
  journal = {arXiv preprint arXiv:2608.31106},
  year    = {2026},
  url     = {https://arxiv.org/abs/2608.31106}
}
\bibliographystyle{ieeenat_fullname}
}

\clearpage
\maketitlesupplementary
\appendix

\section{Implementation Details}
\label{sec:supp_impl}

\subsection{Shared Visual Interface}

The reference backbone is Wan2.2 TI2V-5B.
Its VAE uses 48 latent channels with spatial stride 16, and
the transformer uses $2{\times}2$ spatial patches.
For {Ref}, a 48-channel interpolated LR latent is concatenated
with the noisy target before patch embedding.
The added projection weights are initialized to zero when
expanding a pretrained input layer.
WM uses zero-valued LR channels and its image/camera conditioning;
{Ref} uses a zero image-anchor mask and no camera condition.
The shared transformer and output head serve both roles.

\begin{table}[ht]
\centering
\small
\caption{Reference spatial representations.}
\label{tab:shape_contract}
\begin{tabular}{lccc}
\hline
Role & Pixels & VAE latent & Patch grid \\
\hline
WM & $480{\times}832$ & $30{\times}52$ & $15{\times}26$ \\
{Ref} & $960{\times}1664$ & $60{\times}104$ & $30{\times}52$ \\
\hline
\end{tabular}
\end{table}

Pixel-space degradation uses blur, bicubic downsampling, and
noise while retaining consecutive frames.
Latent conditioning uses nearest-neighbor spatial interpolation
in the reference recipe.
These operations are distinct: bicubic describes the LR video
construction, not the latent upsampler.
Additional latent-condition noise/dropout and a learned latent
upsampler are disabled during reference consolidation.
No {refinement}-specific spatial Swin partition is enabled.

\subsection{Sampling and Temporal State}
\label{sec:supp_role_state}

Table~\ref{tab:role_state} describes the {noisy-forward Ref} consolidation
variant used as the reference for the Method.
{Here Ref denotes high-resolution refinement.}
The shared shift parameter is 5.0; nominal timestep indices
are warped by the common scheduler.
Teacher and student use matching flow/noise parameterizations.
The four-step WM transitions re-noise clean student predictions,
whereas one-step {Ref} has no additional within-chunk query.

\begin{table}[ht]
\centering
\small
\caption{Reference consolidation settings.
Sampling and cache settings are listed separately for each role.}
\label{tab:role_state}
\resizebox{\columnwidth}{!}{
\begin{tabular}{lcc}
\hline
Property & WM & {Ref} \\
\hline
Condition & image, camera, text & LR latent, text \\
Training layout & $[1,4,4,\ldots]$ & $[4,4,\ldots]$ \\
Nominal indices & $[1000,750,500,250]$ & $[1000]$ \\
Temporal span & 25 latent frames & 8 latent frames \\
Rollout & 25 latent frames & 8--16 latent frames \\
Supervised blocks & 6 content blocks & final block \\
Supervised queries & all 4 & sole query \\
History construction & clean recommit & raw noisy-forward KV \\
Temporal RoPE & window reindexing & window reindexing \\
\hline
\end{tabular}}
\end{table}

For clean WM commitment, both networks separately process the
same completed student chunk at context timestep zero.
For {Ref}, each network retains its own noisy-forward entries;
the specialist is also run on unsupervised history blocks.
No student KV tensors are copied into a specialist.
All preceding states and transitions are detached, so the local
velocity loss does not backpropagate through earlier chunks.
All six WM content blocks receive supervision at all four queries.
For {Ref}, a two-to-four-block rollout provides {noisy-forward history}, and only
the final block contributes the regression loss.

\subsection{Temporal RoPE Reindexing}
\label{sec:supp_rope}

Both WM and Ref use local temporal positions within their respective
attention windows. Reindexing addresses positional extrapolation when
inference exceeds the finite temporal spans used for training.
Let $p_i$ and $\widetilde p_i$ be the old and new temporal positions
of a retained token, and let $R(p)$ denote the temporal RoPE rotation.
If $K_i$ already includes the rotation at $p_i$, its updated key is
\begin{equation}
    \widetilde K_i
    =R(\widetilde p_i)R(p_i)^{-1}K_i.
    \label{eq:supp_rope_reindex}
\end{equation}
The corresponding value is retained unchanged.
This updates positional encoding without recomputing the hidden
features of the retained history.
Spatial RoPE coordinates are unaffected.

For Ref with an eight-latent window and four-latent chunks,
the most recent history chunk occupies positions $0$--$3$ and
the incoming chunk occupies $4$--$7$ once the window is full.
On each subsequent shift, the retained keys are reindexed to
$0$--$3$ before the new chunk is processed.
For WM, a retained image anchor occupies the initial position,
and the remaining window contains the most recent content history
and current chunk. The anchor keeps its designated position while
the moving history is reindexed.
The indexing convention is shared between specialist rollouts,
student and teacher histories during TA-OPD, and streaming inference.
Reindexing aligns the range of temporal positions; it does not
make long generated histories identical to training data.

\subsection{Frame Alignment and Startup}

For the causal Wan VAE, $F$ latent frames correspond to
$4(F-1)+1$ pixel frames.
The first latent frame corresponds to the initial observation, while
subsequent four-latent content blocks each produce 16 pixel frames.
WM starts from a one-latent observation anchor. Refinement processes
this anchor before the subsequent content blocks; standalone {refinement}
training uses uniform four-latent blocks.
Noisy-forward refinement retains the KV entries from its generation
pass, whereas clean-recommit refinement adds an HR history forward.

\subsection{Optimization and Specialist Training}

{WM training uses camera-annotated navigation video;
refinement additionally uses natural HR video with synthetic spatial
degradation.}

The causal specialists are trained with DMD after shared-model
preparation. In that stage, the real and learned fake score models
provide distribution-matching guidance to a causal generator;
the resulting generator becomes a TA-OPD teacher.
The reference {refinement} specialist learns with noisy-forward history and
window-local RoPE positions during its DMD rollouts.
Its training includes an auxiliary reconstruction objective,
whereas the subsequent TA-OPD stage uses only local velocity regression.
Specialist preparation and TA-OPD therefore have distinct objectives
while preserving the corresponding generation context policies.

The reference optimizer follows a deterministic WM, WM, {Ref}
update cycle with learning rate $2{\times}10^{-6}$, effective
batch size 16, gradient clipping 0.5, and EMA decay 0.99.
Consolidation uses local flow MSE without an extra reconstruction
or reward term. Both specialists inherit the jointly trained backbone
before task-specific DMD training.

\subsection{Function-Preserving Input Expansion}
\label{sec:supp_zero_concat}

We establish the functional equivalence of the original WM and its
zero-conditioned input expansion with the original parameters fixed.
Let $x$ contain all original patch-embedding input channels and
let $r$ denote the appended LR condition.
For a flattened input patch, the original affine projection is
$P_{W,b}(x)=Wx+b$; the expanded projection is
\begin{equation}
    \widetilde P_{W,A,b}(x,r)=Wx+Ar+b,
    \label{eq:supp_expanded_projection}
\end{equation}
where $A$ contains only the added condition-channel weights.
This notation also represents the implemented 3D convolution,
whose kernel is linear in the input channels.
The expansion copies $W$ and $b$ and initializes $A$ to zero.

\paragraph{Proposition (zero-condition equivalence).}
Suppose the added condition enters only through
Eq.~\ref{eq:supp_expanded_projection}, with no joint
normalization over the original and added channels before projection.
Keep all original weights, conditioning paths, token geometry,
and downstream computations unchanged.
For identical timesteps, WM conditions, and causal states,
setting $r=0$ makes the expanded model's prediction equal to that
of the original model, for any finite $A$, in exact arithmetic.
With $A=0$, the same equality holds for any finite $r$,
provided all other inputs and execution settings remain unchanged.

\paragraph{Proof.}
By linearity,
\begin{equation}
    \widetilde P_{W,A,b}(x,0)=Wx+A0+b=P_{W,b}(x).
    \label{eq:supp_zero_condition}
\end{equation}
Thus both networks receive identical patch embeddings.
Since all remaining operations, parameters, and conditions agree,
their predictions and corresponding cache updates agree.
Starting from equal causal states, induction over denoising queries
and clean-history commitments extends the equality to the WM
rollout when the sampler, random draws, and state-management
policies are identical.  The $A=0$ case follows from $0r=0$.
\hfill$\square$

\paragraph{Gradient implication.}
For a WM loss whose dependence on $A$ is only through this
projection, zero conditioning gives
\begin{equation}
    \nabla_A\mathcal L_{\mathrm{WM}}
    =\sum_p
      \frac{\partial\mathcal L_{\mathrm{WM}}}{\partial h_p}
      r_p^\top
    =0,
    \label{eq:supp_zero_condition_gradient}
\end{equation}
where $p$ indexes projected patches and $h_p$ is their embedding.
This is a statement about the loss gradient, not the optimizer
update: stored momentum or weight decay can still change $A$.

\paragraph{Numerical precision.}
The equalities above hold in exact arithmetic.
Floating-point implementations may differ across kernels;
numerical validation should use matched executions, a declared
tolerance, and isolated WM random streams.

\section{Evaluation Details}
\label{sec:supp_eval}

\subsection{World-Generation Evaluation}

\paragraph{WBench evaluator.}
All WBench results reported in this paper are evaluated with
Doubao as the multimodal judge. We use the same evaluator,
prompt templates, output parsing, dimension definitions, and
aggregation procedure for every method. The reported Average is
the arithmetic mean of the five unrounded WBench dimension scores.

\paragraph{Matched prediction paths.}
The native WM and self-cascade should use identical WM random
draws, conditions, and mutable runtime settings.
Separate per-role random streams prevent {refinement} sampling from changing
later WM noise.
This execution check is distinct from comparing the trained
unified model against the WM specialist.

\begin{figure*}[!t]
\centering
\textbf{Additional Qualitative Results}\par\medskip
\label{sec:supp_qualitative}
Figure~\ref{fig:supp_more_results} shows additional camera-controlled
rollouts and their corresponding high-resolution refinements.
\par\medskip
    \includegraphics[width=\linewidth,height=0.74\textheight,keepaspectratio]{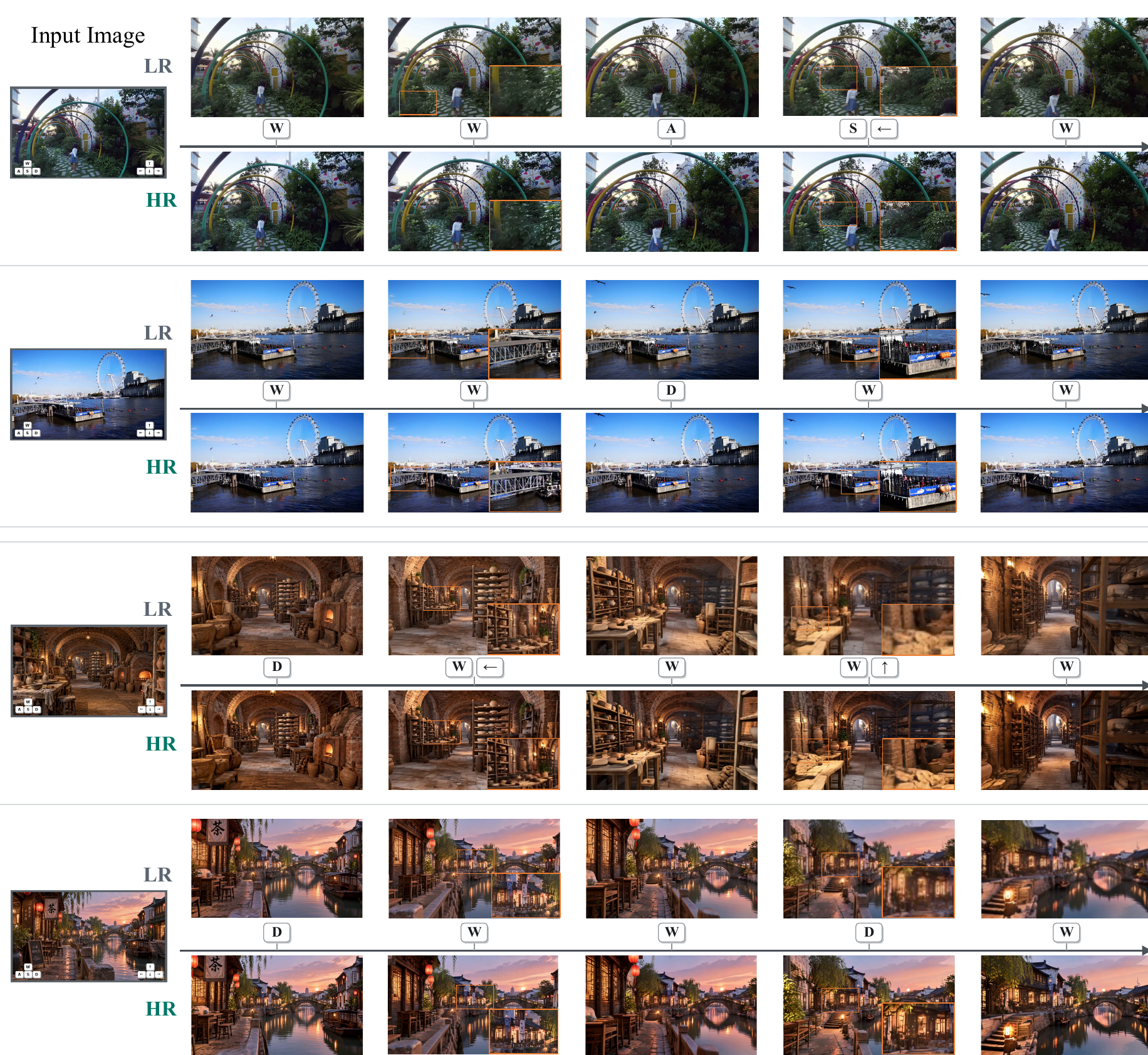}
    \caption{{\textbf{Additional qualitative self-cascade results.}
    Each example shows an initial observation followed by successive
    camera-controlled rollout frames. The \textsc{LR} rows directly
    decode WM-predicted latents with the VAE; the \textsc{HR} rows
    show refinement outputs conditioned on those latents. Key labels
    indicate camera controls, and orange boxes and insets highlight
    matched regions.}}
    \label{fig:supp_more_results}
\end{figure*}

\subsection{Refinement Inputs}
\paragraph{Controlled refinement interface.}
For fixed-input comparisons, the refinement role can receive an
encoded external LR video through the same condition channels.
When comparing pixel-input refiners, all methods receive the same
decoded LR video, which our model re-encodes for conditioning.
Native latent handoff is evaluated separately as part of the
complete self-cascade.

\paragraph{Generated and paired inputs.}
For generated LR, we evaluate refinement without assuming a
unique HR target.
For held-out real-video pairs, PSNR, SSIM, temporal PSNR,
and perceptual distances provide reconstruction diagnostics
\citep{zhang2018lpips,ding2022dists}.
Bicubic upsampling provides a reference for refinement,
alongside content and temporal analysis.

\subsection{Measured Fixed-Input Refinement Protocol}
\label{sec:supp_refinement_runs}
{The two refinement variants in Table~\ref{tab:fixed_lr}
use identical training and evaluation settings and differ only in
refinement-history construction. Both use EMA weights, with the same
initialization, training data, objectives, training-update budget, and
history-window capacity. Clean-recommit KV re-encodes completed HR chunks
at context timestep zero, whereas noisy-forward KV retains the entries
from the conditional noisy generation forward.}

{Table~\ref{tab:fixed_lr} uses the same frozen LR videos
generated by our world model, with matched case IDs,
frame counts, and 24\,FPS timestamps.} The videos contain
193, 289, 385, 481, or 577 frames; all decoded frames are retained.
MUSIQ and CLIP-IQA+ are computed at frames $0,8,16,\ldots$
using \texttt{pyiqa} 0.1.16 (default MUSIQ weights and the
CLIP-IQA+ RN50 learned-prompt model).
We average sampled image-quality scores within each video
and then weight videos equally.

LR-consistency PSNR uses OpenCV AREA downsampling of floating-point
RGB outputs and the mean squared error over all pixels, channels,
and frames of one video, with peak value 1.
For MC L1, we estimate forward and backward Farneb\"ack flow from
the corresponding LR view, with maximum side length 640.
The backward flow warps the preceding HR grayscale frame to the
current frame; forward--backward consistency and image bounds
define a shared validity mask for all methods within a protocol.
The residual excludes a four-pixel HR border. MC HF L1 uses
grayscale minus a Gaussian blur with standard deviation one HR pixel.
These are LR-guided motion-compensated diagnostics, not official
WBench consistency scores.

The full-frame Bicubic baseline uses OpenCV INTER\_CUBIC.
The FlashVSR baseline follows its PIL Bicubic resize-then-crop
implementation; resizing and cropping are performed in memory
without re-encoding the output videos.
Both of our variants and Ultra Flash retain the full frame.
External chunk layouts are not used to infer chunk-boundary measurements.

\paragraph{Interpretation of temporal metrics.}
Motion-aware temporal residuals are interpreted together with
motion magnitude and detail, since static or blurred output
can exhibit low temporal error.
For WBench, official scores retain their original preprocessing;
additional common-resolution and native-crop analyses are
identified separately.

\subsection{Evaluation Accounting and Computational Costs}
\paragraph{Evaluation units.}
Multiple turns or sampling seeds from one case do not count as
independent scenes.

\paragraph{Cost accounting.}
We separate denoising, cache construction, interface conversion,
and VAE decoding.
Peak memory includes both role caches and any resident modules
or preallocated output buffers.
Finite cache windows bound history storage, not necessarily the
memory required when retaining a complete output video.

\section{Additional Experimental Analysis}
\label{sec:supp_ablation}

\subsection{Gradient and Output Diagnostics of Mixed DMD}
\label{sec:supp_gradient_conflict}

We examine historical mixed-DMD runs to characterize task interference.
At each diagnostic step, WM and {Ref} generator gradients are computed
separately at the same model parameters, without an optimizer update
between the two probes. Dot products and squared norms are aggregated
across FSDP shards to obtain global and per-block cosine similarities.
The native probes use each task's sampling policy; their batches and
random draws need not be identical across training steps.

Table~\ref{tab:mixed_dmd_gradients} follows a run initialized from
consistency-distillation step 1200, with WM:{Ref} updates at $3{:}1$,
{Ref} high-noise DMD weight 0.5, and wavelet-reconstruction weight 0.25.
The reported cosines compare the two DMD gradients only.
Negative alignment persists at the sampled steps 600--800 and
extends to all 30 transformer blocks at step 800.
In a separate two-block {Ref} run with reconstruction weight 0.25
and no high-noise reweighting, the step-200 cosine is $-0.1734$
between WM DMD and {Ref} reconstruction, and $-0.1037$ between
WM DMD and the total {Ref} objective. The latter {Ref} gradient has
approximately $3.29$ times the WM gradient norm.

Output diagnostics also show competing quality trends. In a
WM:{Ref}=$1{:}1$ run, mean {Ref} PSNR over 32 validation cases increases
from 26.446 to 27.444 dB between steps 0 and 400. Over 32 WM cases
with matched sample IDs, frame indices, and noise seeds, the median
prediction-to-GT luminance-jump ratio rises from 1.815 to 2.515,
and the corresponding color-jump ratio rises from 1.538 to 2.237.
These ratios measure adjacent-frame changes in global luminance
and color, not comprehensive world consistency. Together, these
diagnostics reveal cross-task gradient interference during mixed
distillation and competing quality trends between prediction and
refinement, supporting separate specialist optimization before
capability consolidation.

\subsection{Reference Specialist Configuration}
\textbf{Specialist teachers for consolidation.}
Table~\ref{tab:teacher_ablation} summarizes the current student
trained with prediction and refinement specialists. Task-specific
SFT supplies score-model supervision for specialist DMD, whose
few-step generators then serve as frozen TA-OPD teachers.
Each teacher constructs its own temporal state under the corresponding
sampling and history policy. The table reports the resulting student's
native WM and fixed-input refinement scores.

\begin{table}[t]
\centering
\small
\caption{Reference student consolidated from two specialist teachers.
WM Inter./Cons.\ are native WBench Interaction/Consistency;
HR Q and TE are fixed-input MUSIQ and MC L1 for noisy-forward refinement.}
\label{tab:teacher_ablation}
\setlength{\tabcolsep}{3pt}
\begin{tabular}{llcccc}
\hline
\shortstack{WM\\teacher} & \shortstack{{Ref}\\teacher} & \shortstack{WM\\Inter.$\uparrow$} & \shortstack{WM\\Cons.$\uparrow$} & HR Q$\uparrow$ & TE$\downarrow$ \\
\hline
$T_{\mathrm{WM}}$ & $T_{{\mathrm{Ref}}}$ & 86.05 & 85.64 & 58.9656 & 0.0212 \\
\hline
\end{tabular}
\end{table}

\subsection{\texorpdfstring{{Refinement-History Ablation}}{Refinement-History Ablation}}
\label{sec:supp_kv_comparison}
\textbf{{Controlled ablation of refinement KV construction.}}
Table~\ref{tab:kv_config} compares clean-recommit KV with the
noisy-forward KV used in Figures~\ref{fig:teaser} and~\ref{fig:pipeline}.
Both variants use TA-OPD and keep WM history clean.
We reuse the fixed-input quality measurements in
Table~\ref{tab:fixed_lr}, obtained on identical frozen LR videos.
{All other training and evaluation settings are held fixed,
isolating the effect of refinement-history construction.}
Clean recommit adds one HR forward per content chunk to construct
history; noisy-forward KV retains entries from the denoising forward.
We report detail, temporal error, and {generator time and FPS}
to measure the quality--cost trade-off.
{Generator time includes HR generation and history
construction before VAE decoding. Clean-recommit and noisy-forward
take 0.60 and 0.30 s per 16-frame content chunk, respectively.
Dividing 16 by these times gives 26.7 and 53.3 generator FPS, yielding
a $2\times$ generator speedup and a 50\% reduction in generator time.}

\subsection{Additional System Metrics}
Table~\ref{tab:supp_hr_system} summarizes system-level WBench scores,
fixed-input refinement quality, and end-to-end generation throughput.
Deployment costs are reported in Table~\ref{tab:deployment}.
\begin{table*}[t]
\centering
\small
\caption{Separate specialist cascade versus unified generation at
$480{\times}832$ LR and $960{\times}1664$ HR.
WBench Avg.\ is the mean of its five official dimensions.
FPS covers end-to-end generation, including WM prediction, refinement,
history construction, and decoding. WBench scores are measured for all
three systems. MUSIQ reports the corresponding refiners
on the fixed decoded LR inputs of Table~\ref{tab:fixed_lr}, rather
than native latent-handoff outputs. Bold marks the highest reported
value in each column.}
\label{tab:supp_hr_system}
\setlength{\tabcolsep}{2.5pt}
\begin{tabular}{lcccccccc}
\hline
System & Avg.$\uparrow$ & Qual.$\uparrow$ & Set.$\uparrow$
& Inter.$\uparrow$ & Cons.$\uparrow$ & Phys.$\uparrow$
& MUSIQ$\uparrow$ & FPS$\uparrow$ \\
\hline
Cascade ($W\rightarrow R$) & {81.58} & {83.72} & {78.21} & 86.80 & \textbf{{87.04}} & \textbf{{72.13}} & \textbf{59.2371} & 10.77 \\
$U\rightarrow U$ (clean-recommit KV) &
{81.53} & \textbf{{83.83}} & {77.81} & 86.98 &
{87.01} & {72.03} & 56.7657 & 9.41 \\
$U\rightarrow U$ (noisy-forward KV) &
\textbf{{81.60}} & 83.43 & \textbf{{78.73}} & \textbf{87.08} &
{86.92} & {71.82} & 58.9656 & \textbf{11.94} \\
\hline
\end{tabular}
\end{table*}

\section{Efficiency and Memory Analysis}
\label{sec:efficiency}

\subsection{Model Storage and Deployment Configuration}
Table~\ref{tab:deployment} compares separate specialists with one
or two resident copies of the unified model under matched hardware,
precision, output size, and batch size.  The two-copy control uses
identical $U$ weights and inference settings to the shared-copy run,
isolating duplicated weight storage from model quality.
{The parameter count in Table~\ref{tab:deployment}
includes all resident DiT copies per rank.}
Backbone parameters, actual resident weight memory, allocated
self-attention KV memory, and full-pipeline peak memory are
reported separately.  Peak memory also includes the VAE,
text encoder, other caches, activations, temporary buffers, and
retained outputs; it is not the sum of separately measured peaks.

\paragraph{{Measurement protocol.}}
{All deployments use four GPUs, BF16, batch size one,
four-step $480{\times}832$ prediction, and one-step $960{\times}1664$
refinement. End-to-end FPS is 16 divided by the median latency of eight
16-frame chunks after six warm-up chunks. The measured latency includes
WM prediction, refinement, history construction, and four-GPU halo-8 MG
decoding. First-chunk latency uses warm kernels and empty streaming
caches. Peak memory is the maximum allocated across GPUs with the
backbone, VAE, text encoder, and caches resident.}

\subsection{KV Storage across Resolutions}
\label{sec:supp_kv_budget}

Prediction and refinement need not retain equally long temporal
windows at the same spatial resolution.
Let $n_L$ and $n_H$ denote the spatial tokens per latent frame,
and let $w$ and $s$ be the allocated WM and {Ref} window capacities
in latent frames, including current-chunk and sink slots.
For a dense self-attention cache with $L$ layers, batch size $B$,
KV width $d_{\mathrm{KV}}$ per layer, and $b$ bytes per element,
one set of key--value states requires $2LBd_{\mathrm{KV}}b$ bytes
per stored token.
The factor two counts keys and values.
We assume identical backbones and cache precision, without cache
sharding, quantization, or token compression.

Writing $\kappa=2LBd_{\mathrm{KV}}b$, the two-role self-cascade
and a direct-HR WM with the same prediction-window capacity have
\begin{align}
    M_{\mathrm{cascade}}^{\mathrm{KV}}
    &=\kappa(n_Lw+n_Hs),\\
    M_{\mathrm{direct}}^{\mathrm{KV}}
    &=\kappa n_Hw.
    \label{eq:supp_kv_direct}
\end{align}
For $\rho=n_H/n_L$, their ratio is
\begin{equation}
    \frac{M_{\mathrm{cascade}}^{\mathrm{KV}}}
         {M_{\mathrm{direct}}^{\mathrm{KV}}}
    =\frac{1}{\rho}+\frac{s}{w}.
    \label{eq:supp_kv_ratio}
\end{equation}
Thus the self-cascade uses less KV storage whenever
$s<(1-1/\rho)w$.
For fixed $s$, each additional WM window slot costs only
$1/\rho$ as much KV storage as a direct-HR slot.
These are storage identities; they do not assume equal prediction
quality at the two resolutions.

\paragraph{Reference configuration.}
For $480{\times}832$ prediction and $960{\times}1664$ refinement,
Table~\ref{tab:shape_contract} gives $n_L=390$, $n_H=1560$,
and $\rho=4$.
With $w=25$ and $s=8$, Eq.~\ref{eq:supp_kv_ratio} is $0.57$,
corresponding to a $43\%$ reduction in self-attention KV payload.

\paragraph{Measured deployment memory.}
Measured deployment costs follow Table~\ref{tab:deployment}:
weight and KV allocations are reported per GPU, while peak memory
is the maximum allocated memory across the four GPUs.
The analytical ratio above describes KV payload, independently of
these measured deployment costs.

The reduction in Eq.~\ref{eq:supp_kv_ratio} comes from retaining
the prediction context at low resolution with a shorter HR
refinement window, and also applies to a separate-model cascade
with the same cache geometry.  Sharing weights additionally
avoids storing a second backbone.
Whether the shorter HR window preserves sufficient detail
consistency is evaluated through refinement and temporal results.

\section{Limitations}
\label{sec:limitations}

Our formulation targets camera-controlled visual prediction,
not an explicit physical state simulator.
High-resolution refinement may introduce unsupported structures
or alter appearance even when it does not feed back into the
native prediction recurrence.
{Although refinement training incorporates synthetic degradations
designed to mimic generation artifacts~\citep{luxury2026ultraflash,zhu2026dreamxcreator},
these degradations may not fully capture the errors in world-model
predictions. A residual gap between training conditions and generated
LR inputs may therefore remain, potentially limiting refinement
robustness during self-cascading.}
Capability retention and temporal robustness must therefore
be interpreted within the evaluated controls, domains, and horizons.

\end{document}